\documentclass[10pt,twocolumn,letterpaper]{article}

\usepackage{cvpr}              
\usepackage[accsupp]{axessibility}  

\definecolor{cvprblue}{rgb}{0.21,0.49,0.74}
\usepackage[pagebackref,breaklinks,colorlinks,allcolors=cvprblue]{hyperref}
\usepackage[normalem]{ulem}
\usepackage{float}
\usepackage{afterpage}

\def\paperID{MEIS-5} 
\def\confName{CVPR}
\def\confYear{2025}

\title{Multi-Agent Systems for Robotic Autonomy with LLMs}

\author{Junhong Chen, Ziqi Yang, Haoyuan G Xu, Dandan Zhang, George Mylonas\thanks{Corresponding author}\\
Imperial College
London\\
{\tt\small
\{junhong.chen16,z.yang21,g.xu23,d.zhang17,george.mylonas\}@imperial.ac.uk}
}

\begin{document}
\maketitle
\begin{abstract}

\vspace{-10pt}
Since the advent of Large Language Models (LLMs), various research based on such models have maintained significant academic attention and impact, especially in AI and robotics. In this paper, we propose a multi-agent framework with LLMs to construct an integrated system for robotic task analysis, mechanical design, and path generation. The framework includes three core agents: Task Analyst, Robot Designer, and Reinforcement Learning Designer. Outputs are formatted as multimodal results, such as code files or technical reports, for stronger understandability and usability. To evaluate generalizability comparatively, we conducted experiments with models from both GPT and DeepSeek. Results demonstrate that the proposed system can design feasible robots with control strategies when appropriate task inputs are provided, exhibiting substantial potential for enhancing the efficiency and accessibility of robotic system development in research and industrial applications.


\end{abstract}

\vspace{-10pt}    
\vspace{-8pt}
\section{Introduction}
\label{sec:intro}
\vspace{-3pt}
\indent In recent years, Large Language Models (LLMs) like OpenAI’s GPT, Meta’s LLaMA, and Anthropic’s Claude, have emerged as powerful tools, enabling advanced language understanding, reasoning, and problem-solving \cite{radford2018improving,touvron2023llama,bai2022constitutional}. They were initially developed for Natural Language Processing (NLP) but now have been integrated into various fields. Particularly in robotics, they have significant potential for promoting robotic autonomy by enhancing perception, control, and decision-making \cite{roy2025gpt,lai2025natural,wang2024llm}. They can translate natural language descriptions into executable robotic actions, reducing the requirements for task-specific programming. Additionally, the rapid growth of LLM-based multi-agent systems (MAS) has demonstrated a strong potential for collaborative tasks by introducing communication mechanisms among multiple agents, which allow robots to share environmental information, adapt strategies, and efficiently work together toward a common goal \cite{liu2024sensingagent,yang2025autohma,zhao2025dual}.

Reinforcement Learning (RL) has witnessed significant advancements in recent years because of its iterative, human-like approach to learning and decision-making. Disparate from traditional data-driven methods, RL typically requires minimal prior knowledge, as agents learn by actively interacting with the environment and receiving feedback through rewards or penalties \cite{mnih2015human,matsuo2022deep}. Such architecture is particularly feasible for sequential decision-making problems in robotics. RL empowers robots to acquire complex behaviors through trial and error, progressively enhancing their performance and allowing more effective adaptability in dynamic or unstructured environments \cite{sun2025multi,yang2022autonomous}. RL has been successfully applied to tasks such as path planning, robot control, object manipulation, and multi-agent coordination, which all represent critical components of achieving robotic autonomy \cite{liu2024robotic,chen2022path}.

Attributed to the powerful language comprehension and execution efficiency of LLMs, their combination could enhance robot learning by integrating language reasoning and decision-making abilities. LLMs can act as advanced planners, converting natural language descriptions into structured, machine-readable commands that guide RL agents to focus on crucial decisions and accelerate policy learning in complex environments or those real cases which lack of control strategies \cite{kheirandish2025llm,du2023guiding,wang2023}. Furthermore, multi-agent communication architectures enable the decomposition of tasks into sub-tasks that can be executed efficiently, either in parallel or sequentially, simplifying control and enhancing efficiency. For instance, by incorporating multi-agent RL, robotic swarms can interpret commands, generate strategies, and dynamically adapt to complex environments based on language-driven patterns and task-activated reasoning from LLMs \cite{liu2024language,lou2024talker}.

In this study, we propose a multi-agent, LLM-based framework to support robotic task analysis, mechanical design, and path generation. The main contributions include:
\begin{itemize}
\item Introduce a modular LLMs-based framework for flexible and scalable robotic development, featuring three core agents for robotic task analysis, robot design, and reinforcement learner training, with sub-agents for extracting codes and summarizing reports from outputs.
    
\item Develop human-friendly methods for generating readable and reliable technical analysis reports, path visualizations, and RL codes, ensuring clarity, practicality, and actionable results for robotic system design.

\item Conduct experiments with several LLMs in different performance levels to validate the generalizability and efficiency of the proposed framework, and ablation studies to clarify the functions and impacts of each agent.

\end{itemize}

\vspace{-6pt}
\section{Related work}
\label{sec:relatedwork}
\vspace{-4pt}

This section will review the application of LLMs and MAS in the field of robotic autonomy, and relevant research on the reinforcement learning based robotic path generation. 




\begin{figure*}[t]
  \centering
   \includegraphics[width=0.85\linewidth]{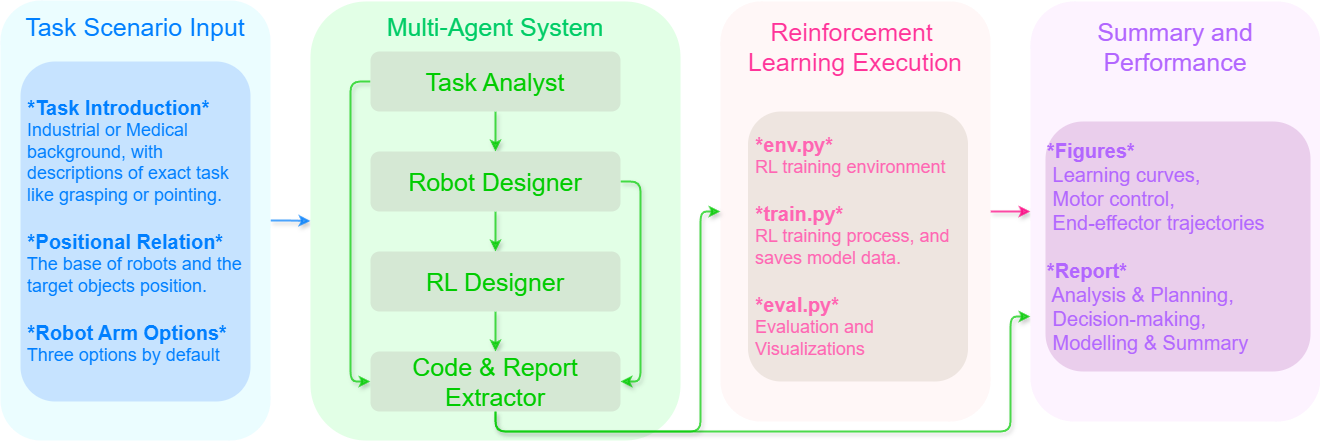}
   \caption{The overview of the proposed multi-agent robot system}
   \label{fig:overview}
   \vspace{-15pt}
\end{figure*}
\vspace{-3pt}
\subsection{LLMs in Robotics}
\vspace{-2pt}
The intelligence required for robotics research has been steadily increasing. Traditional approaches have relied on functional modules with neural network algorithms, some incorporating vision algorithms such as NLP and YOLO to provide a certain degree of intelligence on communication and image reading receptively \cite{wang2024ai,10611534,9376044,dastagir2024nlp}. However, with the widespread adoption of LLMs, robotics research is no longer limited to functionally independent systems but shifts to task-oriented frameworks \cite{9869659,10711245}. Although there is no formal mathematical proof that these models enhance robotic intelligence, recent advancements in task-oriented research based on the capacities of LLMs indeed reveal a promising trend \cite{10802322,rasheed2022review}.

For instance, Google introduced PaLM-E in 2022, a large model specifically designed for robotic task training \cite{driess2023palm}. More recently, LLMs with reasoning capabilities, such as GPT-O1 and DeepSeek-R1, have significantly addressed the challenges of autonomous logical inference in robotics \cite{bottega2025we,peng2025automatic}. LLMs have made significant improvements in various subfields, including autonomous task sequence generation based on task descriptions, detailed task refinement, multi-robot collaboration, and adaptive human-robot interaction \cite{singh2024twostep,10611163,10610065,mandi2024roco,chen2024scalable,kim2024understanding}.
\vspace{-3pt}
\subsection{Multi-Agent Systems in Robotics}
\vspace{-3pt}
Unlike the single robotic module, MAS offers a more human-like and flexible approach to global robotic system design. In MAS-based robotics, each functional module is assigned to a distinct agent, facilitating information processing of different levels and agent-agent communication. By employing a multi-agent framework, robotic systems now walking into a future from a step over single-function demonstrations to fully autonomous systems capable of decision-making, task planning, and execution.

For instance, robots guided by high-level instructions can autonomously determine task execution strategies. The RoCo framework, developed by Mandi, assigns individual robotic agents to each robotic arm in a multi-arm system, enabling coordination through agent interaction \cite{mandi2024roco}. Similarly, multi-agent robots are applicable in multi-robot collaboration systems, where inter-agent communication enhances efficiency and flexibility \cite{li2025large}. MAS improves the adaptability of robotic applications across various subfields and allows them to be seamlessly blended.

\subsection{Reinforcement Learning for Robotic Control}
Reinforcement learning (RL) has become an essential tool in robotics, allowing systems to adapt to diverse task environments and motion constraints. Various RL algorithms are employed based on task-specific requirements. The most commonly used reinforcement learning algorithms include:

\textbf{Q-Learning and DQN (Deep Q-Learning)}: Suitable for tasks requiring discrete iterative motion, such as path finding and obstacle avoidance \cite{zhang2019double,wu2024deep}.

\textbf{A3C (Asynchronous Advantage Actor-Critic)}: Enhances stability and efficiency through asynchronous updates and advantage functions, making it ideal for object grasping \cite{8914201}.

\textbf{PPO (Proximal Policy Optimization) and TRPO (Trust Region Policy Optimization)}: Improve policy optimization techniques, making training for navigation and robotic arm control more stable and efficient \cite{taheri2024deep,quan2023obstacle,8445099}.

\textbf{DDPG (Deep Deterministic Policy Gradient) and SAC (Soft Actor Critic)}: Aim for continuous motion control; SAC further incorporates entropy regularization to enhance exploration capabilities \cite{chen2022path,dang2023rac}.

\textbf{Multi-Agent RL}: Algorithms such as Independent Q-Learning (IQL) and Multi-Agent Deep Deterministic Policy Gradient (MADDPG) enable independent reinforcement learning for each agent, enhancing cooperative intelligence in MAS of robotics \cite{zhuang2024yolo}.
\vspace{-4pt}
\subsection{Path Generation}
\vspace{-2pt}
Path generation is a fundamental problem in robotic kinematics, as it determines how a robot moves at each timestep to complete a given task or reach a target position. Various methods are utilized for trajectory planning:

\textbf{Model Predictive Control (MPC)}: Employs control-based optimization to generate efficient trajectories \cite{7799435,7864379}.

\textbf{Imitation Learning}: Allows robots to mimic expert demonstrations to complete similar tasks \cite{shyam2020imitation}.

\textbf{Reinforcement Learning}: Emphasizes autonomous exploration of feasible trajectories based on environmental feedback, enabling robots to learn and generate paths from nothing \cite{chen2022path,wu2024deep}.

\textbf{Few-Shot and Zero-Shot Learning}: Emerging research focuses on accelerating robot learning for simple tasks with minimal training samples \cite{10869294}.

\textbf{Data-driven Deep Learning}: Enables the robot to generate end-effector trajectories based on sparse constraints or limited task-specific conditions \cite{yang2024automated,zhang2023step}.

These methods, with the advent of LLMs, are showing potential outcomes of robotic task autonomy, enabling robots to perform tasks with less human intervention.

\vspace{-4pt}
\section{Method}
\label{sec:method}

\begin{figure*}[t]
  \centering
   \includegraphics[width=0.95\linewidth]{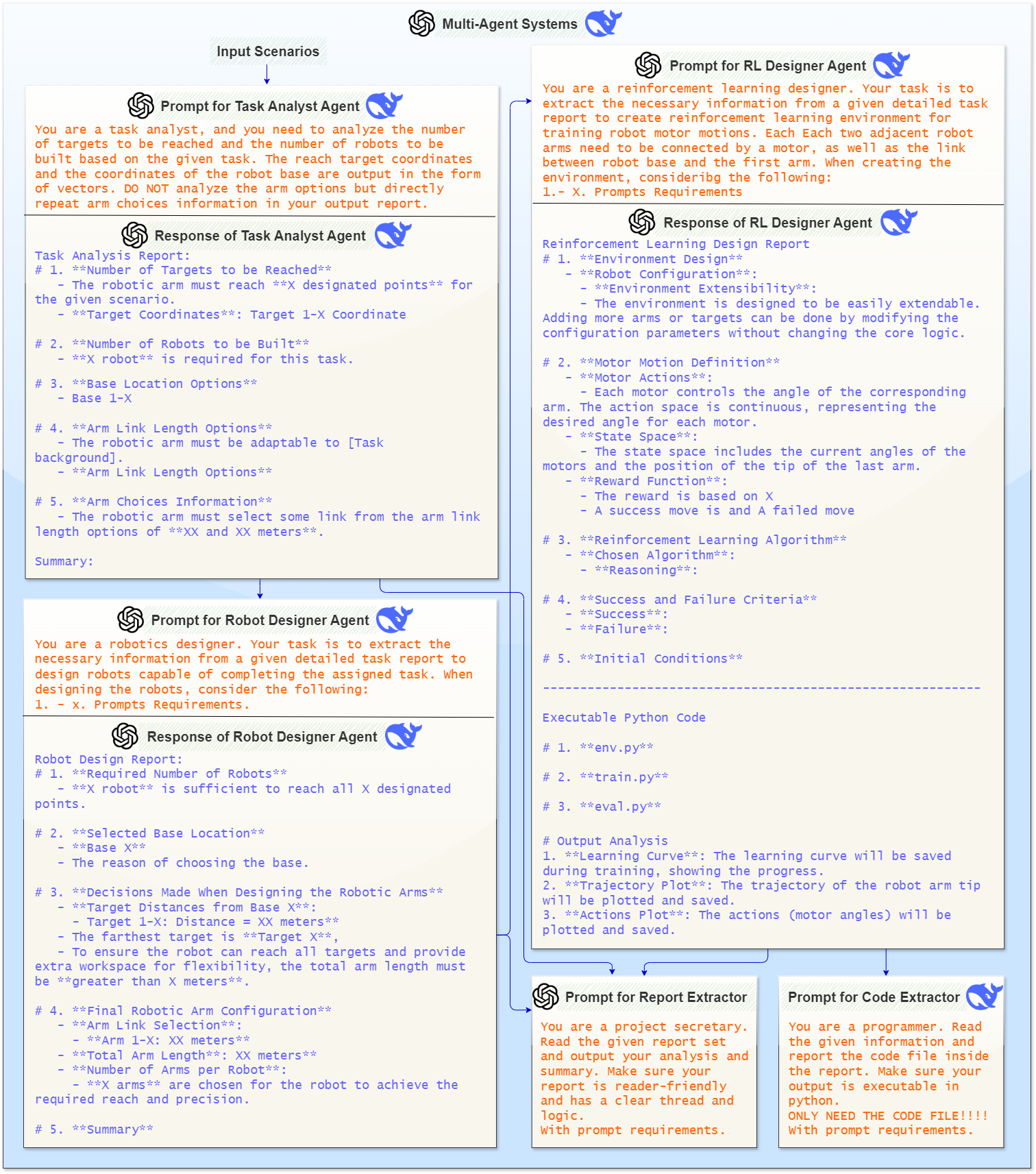}
   \vspace{-8pt}
   \caption{The detailed design of LLMs-based Multi-Agent Systems in the Framework}
   \vspace{-12pt}
   \label{fig:MAS}
\end{figure*}

\vspace{-4pt}
\subsection{Overview of the framework}
\vspace{-2pt}
An overview of the proposed multi-agent robot system framework, is illustrated in Fig.\ref{fig:overview}. The system takes task scenario descriptions as the only input. To ensure system stability, the input must clearly define the following three aspects: 1) the operational environment of the task, 2) the configuration of robot base options and the target positions, and 3) the available robotic arm length options.

Once the task descriptions are clearly specified, the first agent, the Task Analyst, receives the input and performs an engineering-oriented analysis. This agent extracts key information from the task description, establishes a coordinate system based on the provided robot base and target positions, and converts all relevant positional data into the coordinate frame. However, the specification of robotic arm length options is kept without modification. The final output of this agent is a Task Analysis Report, consisting of both extracted and preserved task information.

The second agent, the Robot Designer, processes the Task Analysis Report and extracts necessary details, such as robot base coordinates, target positions, and available arm length options. This agent is responsible for analyzing the task allocation strategy within the given environment, determining the required number of robots, assigning sub-tasks to each robot, and selecting appropriate robotic arms with varying length options. The design process considers both economic efficiency and operational safety by its understanding of the background of task scenarios, ensuring that the selected arms are neither excessively long nor insufficiently short. The final Robotic Design Report is then generated and passed to the next agent.

The third agent, the Reinforcement Learning Designer, utilizes the Robotic Design Report to generate all necessary RL components. This agent is also required to address RL model selection and implementation issues. The final output includes a comprehensive RL implementation report, consisting of an analysis of the RL framework and three key code modules: Environment Definition, Training Script and Evaluation Script, which will be detailed introduced.

Once completing all agent tasks, the reinforcement learning module executes the generated code to train the initial task model. The system then outputs figures and a final report.

\vspace{-2pt}
\subsection{Multi Agents Integration}
\vspace{-2pt}
This section provides a detailed introduction of the multi-agent system architecture, as depicted in Fig.\ref{fig:MAS}. The system consists of three core agents and two additional agents, with the core agents arranged in a linear workflow, while the others dealing with their report for final outputs.

\subsubsection{Task analyst}
The Task Analyst processes the task scenario description as input and performs a structured engineering analysis. The key prompts include:

 1. Determining the number of robots required and assigning base positions based on the task description,
 
 2. Identifying the target points and their coordinates,
 
 3. Summarizing and relaying other task-specific requirements.
 
The Task Analysis Report generated by this agent consists of five key elements:

1. Number of Targets to be Reached, 2. Number of Robots to be Built, 3. Base Location Options, 4. Arm Link Length Options, 5. Arm Choices Information. 

This report is then forwarded to the Robot Designer for further processing.

\subsubsection{Robot designer}
The Robot Designer refines the preliminary engineering analysis into a modeling-ready decision report. The main tasks of this agent include:

1. Extracting key details from the Task Analysis Report to generate a comprehensive system-level plan.

2. Selecting suitable base locations for robots and allocating sub-tasks accordingly.

3. Determining optimal robotic arm configurations, ensuring that each robot can effectively reach all assigned target points while maintaining economic and redundancy considerations.

4. Summarizing all design choices before passing the information to the next agent.

The Robotic System Design Report contains five essential sections: 1. Required Number of Robots, 2. Selected Base Location, 3. Design Decisions for Robotic Arms, 4. Final Robotic Arm Configuration, 5. Summary

Similarly, this report will be the input that is sent to the third core agent: the RL Designer.

\vspace{-2pt}
\subsubsection{RL designer}
\vspace{-2pt}

Since no human demonstrations and prior knowledge is provided, the RL Designer is the most crucial agent of this MAS framework, as it transforms task requirements into an operational reinforcement learning model and generates trajectories for task requirements. However, it deeply relies on the information processed from the Task Analyst and Robot Designer to ensure a well-structured foundation for learning-based training.
This agent has two key aspects:

\textbf{RL Model Selection and Design}: The RL framework is told to be flexible and capable of adaptation. Based on the task environment and objects, the agent needs to select an appropriate RL algorithm and list its reasons behind.

\textbf{RL Code Implementation}: The agent generates RL-related code, defining the environment, motion policies, and success criteria.

The RL Design Report consists of five sections:

1. Environment Design, 2. Motor Motion Definition, 3. Reinforcement Learning Algorithm Selection, 4. Success and Failure Criteria, 5. Initial Conditions

Additionally, three independent code files are generated:

\textbf{Environment Definition (env.py)}: Defines the RL training environment and its initialization, resetting and interaction functions.

\textbf{Training Script (train.py)}: Runs the RL training process, and saves model data.

\textbf{Evaluation Script (eval.py)}: Executes the trained model with given initials and visualizes control data and end-effector trajectories.

\subsubsection{Code and Report extractor}

\textbf{Code Extractor}: Since agents cannot directly execute the generated code, a Code Extractor is necessary to extract all code components mentioned in the reports and divides them into separate files for execution(env.py, train.py, and eval.py).

\textbf{Report Extractor}: Merges all output reports into a final comprehensive report, summarizing the entire intelligent analysis, decision-making, execution strategies, modeling, but excluding code.
\vspace{-2pt}
\subsection{Reinforcement Learning Execution}
\vspace{-2pt}
To keep identical simulation across all experiments, the RL simulations are conducted using OpenAI Gym as the standard environment. This ensures that all scenarios and results are evaluated within a unified benchmarking framework.

Key RL components, states, actions, and awards, are all determined by the RL Designer. This framework highlights the autonomous decision-making capabilities of the MAS, as the self-selected RL algorithms and RL components directly determine the training process and final outcomes.

\vspace{-2pt}
\subsection{Output and Evaluation}
\vspace{-2pt}

When executing the RL scripts (eval.py), the system generates the following outputs:

1. Learning curves from the RL training process.

2. Motor control visualizations, detailing the joint movements of each robot.

3. End-effector trajectories, showing the tip motion paths learned by the robotic arms.

The output figures include: the learning curves, motor control graphs, and robotic end-effector motion trajectories. The final report, generated by the Report Extractor, provides a structured summary of all intelligent analysis, planning, decision-making, modeling, and summary, illustrating the effectiveness of the proposed MAS for RL-powered robotic task autonomy.

The following section will conduct different experiments within this framework, to evaluate its generalizability and key influence of each core agents.

\begin{table*}[h]
    \centering
    \begin{tabular}{|c|lcll|}
\hline
Scenario & \multicolumn{1}{c|}{Task Name}                                                                                                    & \multicolumn{1}{c|}{Base Location Options}                                                                                & \multicolumn{1}{c|}{Target Location}                                                                                              & \multicolumn{1}{c|}{Arm Options}                                                                              \\ \hline
1        & \multicolumn{1}{l|}{Rehabilitation Therapy}                                                                                       & \multicolumn{1}{c|}{(0,0) or (0.5,0)}                                                                                     & \multicolumn{1}{l|}{(0.5,1.2), (0.8,1.5), (1.0,1.0)}                                                                              & 0.8m, 1.0m, 1.2m                                                                                              \\
2        & \multicolumn{1}{l|}{Surgical Instrument Handling}                                                                                 & \multicolumn{1}{c|}{(0,0.5) or (0.2,0.3)}                                                                                 & \multicolumn{1}{l|}{(0.5,0.5), (0.7,0.7), (1.0,0.6)}                                                                              & 0.7m, 0.9m, 1.1m                                                                                              \\
3        & \multicolumn{1}{l|}{Elderly Feeding Assistance}                                                                                   & \multicolumn{1}{c|}{(0,-0.5) or (-0.3,-0.5)}                                                                              & \multicolumn{1}{l|}{(0.4,0.2), (0.5,0.5), (0.6,0.3)}                                                                              & 0.6m, 0.8m, 1.0m                                                                                              \\
4        & \multicolumn{1}{l|}{Physical Therapy Stretching}                                                                                  & \multicolumn{1}{c|}{(0.5,0) or (0.3,-0.2)}                                                                                & \multicolumn{1}{l|}{(0.5,1.0), (0.6,1.2), (0.8,1.1)}                                                                              & 0.9m, 1.1m, 1.3m                                                                                              \\
5        & \multicolumn{1}{l|}{Prosthetic Limb Training}                                                                                     & \multicolumn{1}{c|}{(0,0) or (0.2,-0.2)}                                                                                  & \multicolumn{1}{l|}{(0.3,0.4), (0.5,0.6), (0.7,0.5)}                                                                              & 0.7m, 0.9m, 1.2m                                                                                              \\
6        & \multicolumn{1}{l|}{Assembly Line Placement}                                                                                      & \multicolumn{1}{c|}{(0,0) or (0,0.3)}                                                                                     & \multicolumn{1}{l|}{(0.4,0.3), (0.6,0.5), (0.8,0.4)}                                                                              & 0.8m, 1.0m, 1.2m                                                                                              \\
7        & \multicolumn{1}{l|}{Warehouse Item Sorting}                                                                                       & \multicolumn{1}{c|}{(0,0) or (-0.5,0)}                                                                                    & \multicolumn{1}{l|}{(0.5,1.0), (0.7,1.2), (1.0,1.1)}                                                                              & 0.9m, 1.1m, 1.3m                                                                                              \\
8        & \multicolumn{1}{l|}{Automobile Welding}                                                                                           & \multicolumn{1}{c|}{(0,0) or (1.2,0.5)}                                                                                   & \multicolumn{1}{l|}{(0.4,0.2), (0.6,0.3), (0.8,0.4)}                                                                              & 0.7m, 0.9m, 1.0m                                                                                              \\
9        & \multicolumn{1}{l|}{Pick-and-Place for Electronics}                                                                               & \multicolumn{1}{c|}{(0,0) or (0.2,0.3)}                                                                                   & \multicolumn{1}{l|}{(0.3,0.4), (0.5,0.5), (0.7,0.6)}                                                                              & 0.6m, 0.8m, 1.0m                                                                                              \\
10       & \multicolumn{1}{l|}{Palletizing in Logistics}                                                                                     & \multicolumn{1}{c|}{(0,0) or (0.5,0.5)}                                                                                   & \multicolumn{1}{l|}{(0.4,0.5), (0.6,0.7), (0.8,1.0)}                                                                              & 0.9m, 1.2m, 1.5m                                                                                              \\ \hline
Example  & \multicolumn{4}{l|}{\begin{tabular}[c]{@{}l@{}}"There is a factory with two base for robot manipulators available and the gap of them are 10m, on the front, \\ which is 20m of their middle point, there are 4 boxes need to be picked up, each of them has a 5m gap, \\ and all boxes forms a line which is parallel to the robots base line. The robot could be designed from three \\ following lengths of robot arms: 10m, 5m, and 2m, a robot could have multiple arms to form a serial robot."\end{tabular}} \\ \hline
\end{tabular}
\vspace{-7pt}
    \caption{key details of the ten task scenarios, Scenario 1-5 are industrial cases while Scenario 6-10 are medical cases. Example Description is used as the normal description for the ablation study.}
    \vspace{-8pt}
    \label{tab:10task}
\end{table*}

\begin{table*}[h]
\centering
\begin{tabular}{l|l}
\multicolumn{1}{c|}{One Agent Disable Ablation Condition} & \multicolumn{1}{c}{Two Agents Disable Ablation Condition} \\ \hline
C1: Without Task Analyst (Core Agent 1)                       & C12: Without Task Analyst and Robot Designer (Core Agent 1\&2) \\
C2: Without Robot Designer (Core Agent 2)                     & C13: Without Task Analyst and RL Designer (Core Agent 1\&3)    \\
C3: Without RL Designer (Core Agent 3)                        & C23: Without Robot Designer and RL Designer (Core Agent 2\&3) 
\end{tabular}
\vspace{-7pt}
\caption{Ablation configurations for proposed experiments.}
\vspace{-10pt}
\label{tab:ablation}
\end{table*}

\vspace{-8pt}
\section{Experiments}
\label{sec:experiments}
\vspace{-5pt}

\subsection{Experimental Setup}
Since this study involves four LLMs with varying levels of performance: GPT-4o-mini, DeepSeek-V3, GPT-4o, and DeepSeek-R1. To evaluate the performance of the proposed multi-agent system with different models, we conducted two types of experiments to assess whether these LLMs function well or not within the framework:

\textbf{Generalization Across Tasks}: This experiment evaluates how well when different AI models are adapted to different task descriptions within the MAS. 

\textbf{Ablation Study}: This experiment quantifies the impact of three core agents in the MAS by disabling one or two agents separately to evaluate their output performance.
\vspace{-4pt}
\subsection{Generalization Across Tasks}
\vspace{-4pt}
To evaluate the generalization capability of both the models and the multi-agent framework, the paper designed ten task scenarios, divided into two categories: five industrial tasks and five medical tasks. The dual-scenario design may force the models to face different realistic case-specific challenges, leading to a generalizability check of task analysis of how different AI models perform. Each AI model is required to process all ten task descriptions and execute them using the proposed multi-agent system.

\textbf{Scenario Design:} Table.\ref{tab:10task} presents key details of the ten task scenarios. Recall that the input must contain:

Task title and description

Robot and target object positional information

Options for robotic arm lengths

To rigorously test the comprehension and summarization capabilities of the models, each task description consists of 100 to 150 words detailing the task requirements. Table.\ref{tab:10task} also provides an example task description at the bottom.

\vspace{-4pt}
\subsection{Ablation Study}
\vspace{-2pt}
To systematically quantify the role of each core agent within the framework and to evaluate the framework’s sensitivity to the level of details of task descriptions, we conduct the ablation study from two perspectives:

\subsubsection{Robustness to Task Description}
To investigate how the system responds to different detail levels of task description, we conduct ablation tests using three levels of task description length ranging from highly abstract instructions (e.g., “Pick up a box”) to normal(as the example in Table.\ref{tab:10task}), and to highly detailed specifications (e.g., “In a medical environment, grasping surgical tools requires extremely precise handling. For the current target, a medical-grade surgical instrument, previous studies typically utilize a three-joint robotic arm. The grasping process should approach the object at a stable yet relatively high speed before transitioning into a fine-tuned gripping maneuver.”).

\begin{table*}[htbp]
\centering
\resizebox{\textwidth}{!}{%
    \begin{tabular}{c}
        \includegraphics{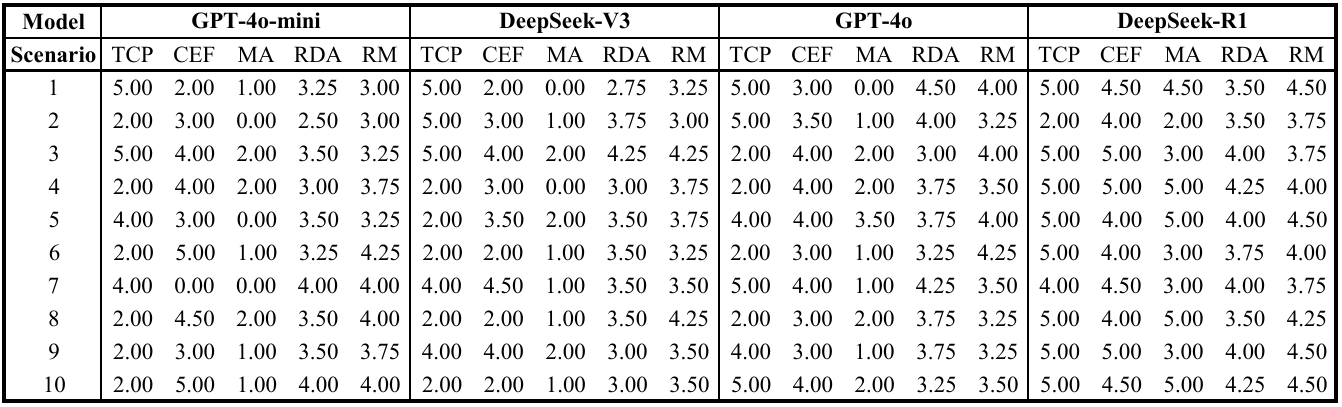}
    \end{tabular}%
}
\vspace{-10pt}
\caption{The score of five evaluation metrics across ten task descriptions. Metrics are: TCP-Task Completion Progress; CEF-Code Execution Feasibility; MA-Model Alignment; RDA-Robot Design Adaptability; RM-Report Maturity}
\label{tab:general}
\vspace{-12pt}
\end{table*}
\subsubsection{Impact of Key Agents}

\begin{table}[htbp]
\centering
\resizebox{\columnwidth}{!}{%
    \begin{tabular}{c}
        \includegraphics[width=1\linewidth]{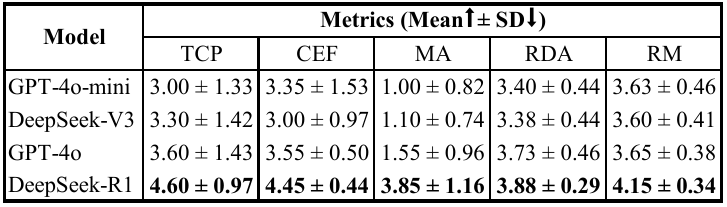}
    \end{tabular}%
}
\vspace{-10pt}
\caption{The Mean and SD values of scores of five evaluation metrics across ten task scenarios on each model.}
\label{tab:general_stat}
\vspace{-15pt}
\end{table}

To quantify the contribution of each core agent, this time we perform controlled experiments using a fixed task description (Example in Table.\ref{tab:10task}). During this experiment, we systematically disable specific one or two agents while ensuring all other components function normally. Table.\ref{tab:ablation} outlines the specific ablation configurations.
\vspace{-2pt}
\subsection{Evaluation Metrics}
\vspace{-2pt}
To quantify those performances, we define the following evaluation metrics for both experiments, each metric has a score ranging from 0 to 5:

\textbf{Task Completion Progress (TCP)}: Measures to what extent each task successfully and properly progresses through the multi-agent system. Due to the purpose of ablation studies, the agent may not finish all of those cases.

\textbf{Code Execution Feasibility (CEF)}: Assesses whether the generated code runs successfully. LLMs may not always have correct and executable code. Some code will run smoothly after minor fixes, but some will generate code that does not run smoothly or use the correct libraries, function calls, and the correct version.

\textbf{Model Alignment (MA)}: Determines if the final execution outcomes meet task requirements based on the generated code. While code could be run properly, the result of each case may not meet the initial requirements or even without convergence.

\textbf{Robot Design Adaptability (RDA)}: Evaluates whether the robot design decisions align with task requirements. Robot design requires years of field operation and design experience, robot may not be designed well with all experimental cases.

\textbf{Report Maturity (RM)}: Evaluates the clarity and completeness of the final output report, ensuring it provides a structured analysis of each step. The final report reflects the MAS' global cognition and detail processing ability for the given task description, and the level of this ability varies with different models or cases.

While TCP, CEF, and MA are objective metrics, the other two metrics, RDA and RM, are relatively subjective, since the primary purpose of the multi-agent system is to enhance intelligent decision-making. To ensure a convincible evaluation, four researchers with several years of research specializing in robotics and LLMs independently score these subjective metrics, and the average of the four scores for these two metrics is used as the final score.

\vspace{-7pt}
\section{Results}
\label{sec:results}
\vspace{-5pt}
The experimental results for the experiments discussed in the previous sections are presented separately in the following two subsections. In addition, an example of output figures from code execution is shown in the last subsection.

\begin{table*}[htbp]
\centering
\resizebox{\textwidth}{!}{%
    \begin{tabular}{c}
        \includegraphics{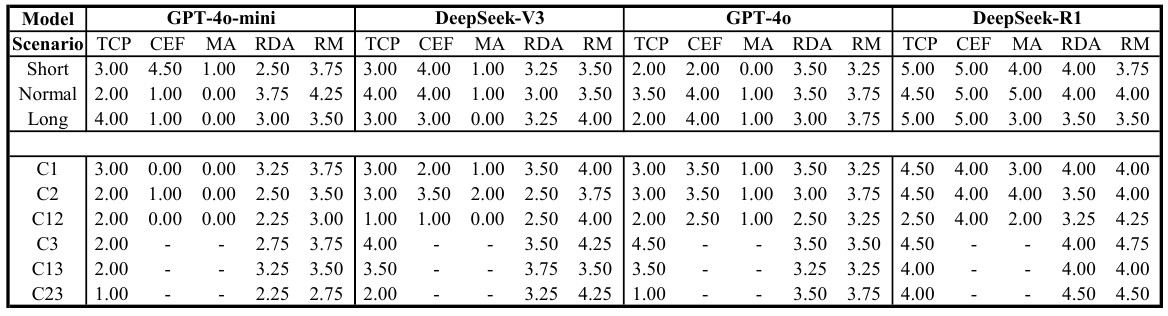}
    \end{tabular}%
}
\vspace{-10pt}
\caption{The score of five evaluation metrics for the ablation study. Metrics are: TCP-Task Completion Progress; CEF-Code Execution Feasibility; MA-Model Alignment; RDA-Robot Design Adaptability; RM-Report Maturity. Short, Normal, and Long indicate the task description length. C1, C2, etc. are the numbered ablation conditions same as Table.\ref{tab:ablation}}
\vspace{-10pt}
\label{tab:ablaresults}
\end{table*}
\vspace{-4pt}
\subsection{Generalization Across Tasks}
\vspace{-2pt}

The score of evaluation metrics for generalization capability is shown in Table.\ref{tab:general} and Table.\ref{tab:general_stat}. Among the four models evaluated in this study, DeepSeek-R1 outperformed the others across all five evaluation metrics.

In TCP, one of the first three objective metrics, GPT-4o-mini and DeepSeek-V3 occasionally achieved high scores. However, their low scores were primarily attributed to two facts: 1) generate information that was not exist in the task requirements; 2) frequent decision-making errors, particularly in the robotic arm design. A common issue was their difficulty in correctly applying mathematical calculations, leading to systematic design errors. GPT-4o also encountered these problems, but less frequently, resulting in an overall better performance. DeepSeek-R1 performed the best due to its self-correction capabilities (reasoning model). This allowed it to verify and refine its decision-making process throughout task execution. Differing from other models, its occasional errors arose when making decisions between optimal and suboptimal solutions for robot design.

Regarding CEF, all models except DeepSeek-R1 received scores in the range of approximately 3. Common issues are due to inconsistent library and function calls, variations in variable formats, and missing statements in code generation. While discrepancies in library versions and function calls can be addressed with minor adjustments, missing statements or wrong code structure usually result in bad outcomes, and show their limitations in generating executable code.

MA scores for the first three models were relatively low, causing their SD also looks relatively small. A key issue was that once the former analysis of the output report contained errors, then subsequent code execution deviated from the task requirements, the system completely got lost. Similarly, reinforcement learning procedures based on error design failed to converge, resulting in no meaningful results. Even DeepSeek-R1 occasionally came up with this issue. However, in most cases, it successfully generated the required output and ended in a convergent solution. An interesting phenomenon was that, among all 40 execution cases, DeepSeek-R1 was the only model that showed variability in selecting RL algorithms, between SAC and PPO. The other models always used PPO for these tasks.

For one of two subjective metrics, RDA, the overall scores of all models were relatively similar. This is likely due to the nature of LLMs in text generation. However, the reasons behind each models varied: GPT-4o-mini occasionally generated text that looking good at first glance, but finding errors when looking into details; DeepSeek-V3 exhibited inconsistencies in report quality; GPT-4o performed slightly better, with output quality close to DeepSeek-R1; DeepSeek-R1 did generate good reports, but sometimes it listed too much details that unable to find the key points.

Finally, in terms of RM, the first three models demonstrated relatively similar performance, while DeepSeek-R1 achieved a slightly higher score. This indicates that the DeepSeek-R1 may have better capability of generating proficient reports.

In summary, the approximate model performance level under this framework is as follows: $\text{GPT-4o-mini} \approx \text{DeepSeek-V3} < \text{GPT-4o} < \text{DeepSeek-R1}$.

\subsection{Ablation Study}
\vspace{-2pt}
In Table.\ref{tab:ablaresults}, for the first ablation study examining the impact of input task description length, the results indicate that more extensive task details do not correlatively enhance output performance in a MAS. The optimal description length leads to effective outputs in most cases, as reflected in the "Normal" column. We believed that excessively long input description lightens the weight of key information inside the output, while overly simple descriptions fail to provide sufficient contextual relevance, particularly for robot design and code generation.

It is likely that both 'Long' and 'Short' input introduce instability into the system. A proper input length enhances the creativity of the overall report while preventing task requirements from becoming biased due to insufficient information. Additionally, CEF and MA exhibit significant fluctuations in models other than DeepSeek-R1. While the code may still execute with minor modifications, decision-making in task analysis and reinforcement learning remains important. Additionally, DeepSeek-R1 also shows deviations in final outputs due to fluctuations in input description length. However, the two subjective metrics show similar values, ranging between 3 and 4. This may suggest that despite increased task difficulty, the model's outputs appear similar for the task report. The key differences lie in the process of transforming reports into actual designs, code, and results. 

Regarding the second ablation study, in the absence of Core Agent 1, the generated results exhibit significant instability. Compared to the "Normal" column in the previous experiment, Task analyst plays a crucial role in defining the overall task direction. Without its guidance, all subsequent generation processes become ambiguous. Even DeepSeek-R1 experiences a decline in performance metrics when losing a high-level analytical summary.
In the experimental comparison of Core Agent 2, Robot Designer, the system’s information processing stopped due to its absence. RL designer struggles to obtain sufficient data for modeling and learning, as meaningful results become difficult to generate. While DeepSeek-R1 shows an advantage by autonomously supplementing missing information, its accuracy and relevance remain uncertain. An interesting observation is that for GPT-4o-mini, the absence of Robot Designer leads to some blank space in reinforcement learning code generation. This causes its code to be entirely non-executable without manual intervention. 
At last, the impact of Core Agent 3, RL designer, is obvious, no reinforcement learning or control logic is present without RL Designer, and the tasks remain confined solely to the robot design stage.

\vspace{-8pt}
\section{Conclusion}
\label{sec:conclusion}
\vspace{-5pt}
In this paper, we proposed a MAS framework for robotic autonomy with LLMs. The proposed multi-agent framework can effectively design feasible robot configurations and produce corresponding control solutions when the task requirements are provided through natural language prompts with proper details. For LLMs with knowledge and capabilities at different levels, lower ones, especially GPT-4o-mini, and DeepSeek-v3, the robot design outcomes are not always feasible, resulting in wrong RL design while GPT-4o meets the borderline of requirements. The reasoning model DeepSeek-R1 keeps a higher quality output with quite a few mistakes. This approach exhibits promising potential for enhancing the efficiency and accessibility of robotic system development based on the knowledge of LLMs instead of human-only strategies, offering valuable insights for future advancements in intelligent robotics and industrial applications.
Future work could focus on the following aspects: introducing obstacles and dynamic objects in the task scenarios; extending the proposed system into a multi-level, nested hierarchical interactive agent architecture; integrating large vision-language models (LvLMs) to extract spatial information from real-world scenarios; and deploying a voice module for continuous human-machine natural language communication.

{
    \small
    \bibliographystyle{unsrt}
    \bibliography{main}
}


\end{document}